\documentclass[10pt]{article}

\usepackage{psfig}
\usepackage{graphicx}
\usepackage{harvard}

\newcommand{\spc}[1]{\rule{#1}{0mm}}
\newcommand{\vspc}[1]{\rule{0mm}{#1}}
\newcommand{\tabhdr}[2]{\makebox[#1][r]{#2}}
\newcommand{\aList}[1]{\langle{#1}\rangle}	
\newcommand{\aPair}[2]{\langle\,{#1}\,,\;{#2}\,\rangle}	
\newcommand{\fstLabel}[1]{\small{\tt{#1}}}
\newcommand{\pTa}[1]{P_{T_1}({#1})}	
\newcommand{\pTb}[1]{P_{T_2}({#1})}	

\newcommand{\vecvv}[2]
{\left[\begin{array}{c}{#1}\\ {#2}\end{array}\right]}

\newcommand{\vecvvv}[3]
{\left[\begin{array}{c}{#1}\\ {#2}\\ {#3}\end{array}\right]}

\newcommand{\vecvvvv}[4]
{\left[\begin{array}{c}{#1}\\ {#2}\\ {#3}\\ {#4}\end{array}\right]}

\begin{document}

\title{Part-of-Speech Tagging with Two Sequential Transducers}
\author{Andr\'e Kempe}
\date{Xerox Research Centre Europe}

\maketitle

\begin{abstract}
The article presents a method of constructing and applying a cascade
consisting of a left- and a right-sequential finite-state transducer,
$T_1$ and $T_2$,
for part-of-speech disambiguation.
In the process of POS tagging,
every word is first assigned a unique ambiguity class
that represents the set of alternative tags
that this word can occur with.
The sequence of the ambiguity classes of all words of one sentence
is then mapped by $T_1$ to a sequence of reduced ambiguity classes
where some of the less likely tags are removed.
That sequence is finally mapped by $T_2$ to a sequence of single tags.
Compared to a Hidden Markov model tagger,
this transducer cascade has the advantage
of significantly higher processing speed,
but at the cost of slightly lower accuracy.
Applications such as Information Retrieval,
where the speed can be more important than accuracy,
could benefit from this approach.
\end{abstract}

\section{Introduction}

We present a method of constructing and applying
a cascade consisting of a left- and a right-sequential
finite-state transducer (FST), $T_1$ and $T_2$,
for part-of-speech (POS) disambiguation.

\begin{figure}[h]
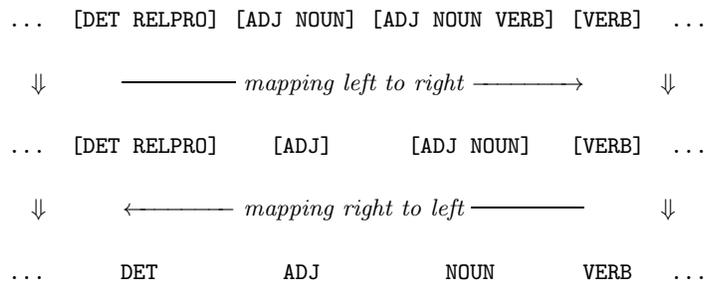
  
\begin{center}
\small
\begin{minipage}{110mm}
\begin{verbatim}
...  [DET RELPRO] [ADJ NOUN] [ADJ NOUN VERB] [VERB]  ...
\end{verbatim}

\vspace{1mm}
\spc{2mm} $\Downarrow$ \spc{8mm}
\rule[1mm]{15mm}{0.1mm}
{\small \it mapping left to right}
\makebox[15mm]{\rightarrowfill}
\spc{8mm} $\Downarrow$

\vspace{1mm}
\begin{verbatim}
...  [DET RELPRO]    [ADJ]      [ADJ NOUN]   [VERB]  ...
\end{verbatim}

\vspace{1mm}
\spc{2mm} $\Downarrow$ \spc{8mm}
\makebox[15mm]{\leftarrowfill}
{\small \it mapping right to left}
\rule[1mm]{15mm}{0.1mm}
\spc{8mm} $\Downarrow$

\vspace{1mm}
\begin{verbatim}
...      DET          ADJ          NOUN       VERB   ...
\end{verbatim}
\end{minipage}

\vspace{4mm}
\caption{
Part of an input, an intermediate, and an output sequence
in the FST cascade (example)
	\label{f-sequences}}
\normalsize
\end{center}
\end{figure}

In the process of POS tagging,
we first assign every word of a sentence a unique ambiguity class $c_i$
that can be looked up in a lexicon encoded by a sequential FST.
Every $c_i$ is denoted by a single symbol,
e.g. ``\fstLabel{[ADJ~NOUN]}'',
although it represents a set of alternative tags
that a given word can occur with.
The sequence of the $c_i$ of all words of one sentence
is the input to our FST cascade
(Fig.~\ref{f-sequences}).
It is mapped by $T_1$, from left to right,
to a sequence of reduced ambiguity classes $r_i$.
Every $r_i$ is denoted by a single symbol,
although it represents a set of alternative tags.
Intuitively,
$T_1$ eliminates the less likely tags from $c_i$,
thus creating $r_i$.
Finally,
$T_2$ maps the sequence of $r_i$, from right to left,
to an output sequence of single POS tags $t_i$.
Intuitively,
$T_2$ selects the most likely $t_i$ from every $r_i$
(Fig.~\ref{f-sequences}).

Compared to a Hidden Markov model (HMM)
\cite{rabiner},
this FST cascade has the advantage of significantly higher processing speed,
but at the cost of slightly lower accuracy.
Applications such as Information Retrieval,
where the speed can be more important than accuracy,
could benefit from this approach.

Although our approach is related to the concept of bimachines
\cite{schutzenberger61}
and factorization
\cite{elgot},
we proceed differently in that we build two sequential FSTs
directly and not by factorization.

This article is structured as follows.
Section~\ref{s-classes}
describes how the ambiguity classes and reduced ambiguity classes
are defined based on a lexicon and a training corpus.
Then, Section~\ref{s-probs}
explains how the probabilities of these classes in the context
of other classes are calculated.
The construction of $T_1$ and $T_2$ is shown in
Section~\ref{s-construct}.
It makes use of the previously defined classes and their probabilities.
Section~\ref{s-applic}
describes the application of the FSTs to an input text,
and Section~\ref{s-results}
finally compares the FSTs to an HMM tagger,
based on experimental data.

\section{Definition of Classes \label{s-classes}}

Instead of dealing with lexical probabilities of individual words
\cite{church},
many POS taggers group words into {\it ambiguity classes}\/
and deal with lexical probabilities of these classes
\cite{cutting,kupiec92}.
Every word belongs to one ambiguity class
that is described by the set of all POS tags that the word can occur with.
For example,
the class described by {\tt \{NOUN,VERB\}} includes all words
that could be analyzed either as noun or verb
depending on the context.
We follow this approach.

Some approaches make a more fine-grained word classification
\cite{daelemans,tzoukermann}.
Words that occur with the same alternative tags,
e.g., {\tt NOUN} and {\tt VERB},
can here be assigned different ambiguity classes
depending on whether they occur
more frequently with one or with the other tag.
Although this has proven to increase the accuracy
of HMM-based POS disambiguation,
it did not significantly improve our method.
After some investigations in this direction,
we decided to follow the simpler classification above. \\

Before we can build the FST cascade,
we have to define ambiguity classes,
that will constitute the input alphabet of $T_1$,
and {\it reduced ambiguity classes}\/,
that will form the intermediate alphabet of the cascade,
i.e., the output of $T_1$ and the input of $T_2$.

Ambiguity classes $c_i$ are defined from the training corpus and lexicon,
and are each described by a pair consisting of
a tag list $\hat t(c_i)$ and a probability vector $\vec p(c_i)\,$:
\begin{equation}
\hat t(c_i) = \aList{t_{i1},t_{i2},...,t_{i,n}}		\spc{10ex}
\vec p(c_i) = \vecvvvv{p(t_{i1}|c_i)}{p(t_{i2}|c_i)}{\vdots}{p(t_{i,n}|c_i)}
    \label{e-ac}
\end{equation}

\noindent
For example:
\begin{equation}
\hat t(c_1) = \aList{{\tt ADJ},{\tt NOUN},{\tt VERB}}	\spc{10ex}
\vec p(c_1) = \vecvvv{0.29}{0.60}{0.11}
    \label{e-ac-exm1}
\end{equation}

\noindent
which means that the words that belong to $c_1$
are tagged as {\tt ADJ} in 29~\%,
as {\tt NOUN} in 60~\%,
and as {\tt VERB} in 11~\% of all cases in the training corpus.

When all $c_i$ are defined,
a class-based lexicon,
that maps every word to a single class symbol,
is constructed from the original tag-based lexicon,
that maps every word to a set of alternative tag symbols.
In the class-based lexicon,
the above $c_1$
(Eq.~\ref{e-ac-exm1})
could be represented, e.g., by the symbol
``\fstLabel{[ADJ NOUN VERB]}''.  \\

We describe a reduced ambiguity classes $r_i$ also by a pair consisting of
a tag list $\hat t(r_i)$ and a probability vector $\vec p(r_i)\,$.
Intuitively,
an $r_i$ can be seen as a $c_i$
where some of the less likely tags have been removed.
Since at this point we cannot decide
which tags are less likely,
all possible subclasses of all $c_i$ are considered.
To generate a complete set of $r_i$,
all $c_i$ are split
into all possible subclasses $s_{ij}$ that are assigned
a tag list $\hat t(s_{ij})$
 containing a subset of the tags of $\hat t(c_i)$,
and an (un-normalized) probability vector $\vec p(s_{ij})$
 containing only the relevant elements of $\vec p(c_i)\,$.
For example,
the above $c_1$
(Eq.~\ref{e-ac-exm1})
is split into seven subclasses $s_{1j}$~:
\begin{eqnarray}
\hat t(s_{1,0}) \; = &\! \aList{{\tt ADJ},{\tt NOUN},{\tt VERB}}
  \spc{8ex}
  \vec p(s_{1,0}) &\! = \; \vecvvv{0.29}{0.60}{0.11}
    			\nonumber	\\
\hat t(s_{1,1}) \; = &\! \aList{{\tt NOUN},{\tt VERB}}
  \spc{13ex}
  \vec p(s_{1,1}) &\! = \; \vecvv{0.60}{0.11}
    \label{e-rc-exm}				\\
\hat t(s_{1,2}) \; = &\! \aList{{\tt ADJ},{\tt VERB}}
  \spc{14ex}
  \vec p(s_{1,2}) &\! = \; \vecvv{0.29}{0.11}
    			\nonumber	\\
{\mit etc.}	\spc{8ex} & &	\nonumber
\end{eqnarray}

Different $c_i$ can produce a $s_{ij}$
with the same tag list $\hat t(s_{ij})$
but with different probability vectors $\vec p(s_{ij})\,$;
e.g., the classes with the tag lists
$\aList{{\tt ADJ},{\tt NOUN},{\tt VERB}}$,
$\aList{{\tt NOUN},{\tt VERB}}$, and
$\aList{{\tt ADJ},{\tt ADV},{\tt NOUN},{\tt VERB}}$
can all produce a subclass with the tag list
$\aList{{\tt NOUN},{\tt VERB}}\,$.
To reduce the total number of subclasses,
all $s_{ij}$ with the same tag list $\hat t(s_{ij})$ are clustered,
based on the {\it centroid method}
\cite[p.~136]{romesburg},
using the {\it vector cosine} as the similarity measure between clusters
\cite[p.~201]{salton}.
Each final cluster constitutes a reduced ambiguity class $r_y$.
If we obtain, e.g., three $r_y$ with the same tag list
$\hat t(r_y)\!=\!\aList{{\tt NOUN},{\tt VERB}}$
but with different (re-normalized) probability vectors:
\begin{equation}
\vec p(r_1) = \vecvv{0.89}{0.11}	\spc{5ex}
\vec p(r_2) = \vecvv{0.57}{0.43}	\spc{5ex}
\vec p(r_3) = \vecvv{0.09}{0.91}
	\label{e-ri-probvec-exm}
\end{equation}

\noindent
we represent them in an FST by three different symbols, e.g.,
``\fstLabel{[NOUN VERB]\_R\_1}'',
``\fstLabel{[NOUN VERB]\_R\_2}'', and
``\fstLabel{[NOUN VERB]\_R\_3}''.

\section{Contextual Probabilities \label{s-probs}}

$T_1$ will map a sequence of $c_i$, from left to right,
to a sequence of $r_i$.
Therefore,
the construction of $T_1$ requires estimating the most likely $r_i$
in the context of both the current $c_i$ and the previous $r_{i-1}$
(wrt. the current position $i$ in a sequence).
To determine this $r_i$,
a probability $\pTa{t_{ij}}$
is estimated for every POS tag $t_{ij}$ in $c_i\,$.
In the initial position,
$\pTa{t_{ij}}$ depends on the preceding sentence boundary $\#_{i-1}$
and the current $c_i$
which are assumed to be mutually independent:
\begin{eqnarray}
\pTa{t_{ij}}  & = & p(t_{ij} | \#_{i-1}\; c_i)		\nonumber \\
\vspc{1ex}				\nonumber  \\
  & = &    \frac{p(t_{ij}\; \#_{i-1}\; c_i)}
		{p(\#_{i-1}\; c_i)}			\nonumber \\
\vspc{1ex}				\nonumber  \\
  & = &    \frac{p(\#_{i-1}\; c_i | t_{ij}) \cdot p(t_{ij})}
		{p(\#_{i-1}\; c_i)}			\nonumber \\
\vspc{1ex}				\nonumber  \\
  & \approx &
	   \frac{p(\#_{i-1} | t_{ij}) \cdot p(c_i | t_{ij}) \cdot p(t_{ij})}
		{p(\#_{i-1}) \cdot p(c_i)}		\nonumber \\
\vspc{1ex}				\nonumber  \\
  & = &    \frac{\frac{p(\#_{i-1}\; t_{ij})}{p(t_{ij})}
		 \cdot \frac{p(c_i\; t_{ij})}{p(t_{ij})}
		 \cdot p(t_{ij})}
		{p(\#_{i-1}) \cdot p(c_i)}		\nonumber \\
\vspc{1ex}				\nonumber  \\
  & = &    \frac{p(t_{ij}\; \#_{i-1}) \cdot p(t_{ij}\; c_i)}
		{p(t_{ij}) \cdot  p(\#_{i-1}) \cdot p(c_i)}	\nonumber \\
\vspc{1ex}				\nonumber  \\
  & = &	   \frac{p(t_{ij} | \#_{i-1}) \cdot p(t_{ij} | c_i)}
		{p(t_{ij})}
  \label{e-pT1-init}
\end{eqnarray}

\noindent
The latter $p(t_{ij}|c_i)$ can be extracted
from the probability vector $\vec p(c_i)$,
and $p(t_{ij}|\#_{i-1})$ and $p(t_{ij})$
can be estimated from the training corpus.

In another than the initial position,
$\pTa{t_{ij}}$ depends on
the preceding $r_{i-1}$ and the current $c_i$
which are assumed to be mutually independent:
\begin{equation}
\pTa{t_{ij}} = p (t_{ij} | r_{i-1}\; c_i)
\approx \frac{p (t_{ij} | r_{i-1}) \cdot p (t_{ij} | c_i)}
		{p (t_{ij})}
  \label{e-pT1-normal}
\end{equation}

\noindent
The latter $p(t_{ij}|r_{i-1})$ is estimated by:
\begin{eqnarray}
p (t_{ij} | r_{i-1}) & = &
\sum\limits_k p (t_{ij} | t_{i-1,k}) \cdot p (t_{i-1,k} | r_{i-1})
  \label{e-aLR-T1}	\\
  & ~ & \spc{10ex}{\rm with}\spc{3ex}
    t_{ij} \in \hat t(c_i) \; ; \;\; t_{i-1,k} \in \hat t(r_{i-1}) \nonumber
\end{eqnarray}

\noindent
where $p(t_{ij}|t_{i-1,k})$ can be estimated from the training corpus,
and $p(t_{i-1,k}|r_{i-1})$ can be extracted
from the probability vector $\vec p(r_{i-1})$
of the preceding $r_{i-1}$.

To evaluate all tags of the current $c_i\,$,
a list ${\hat{\cal P}}(c_i)$
containing pairs $\aPair{t_{ij}}{\pTa{t_{ij}}}$
of all tags $t_{ij}$ of $c_i$
with their probabilities $\pTa{t_{ij}}$
(Eq.s~\ref{e-pT1-init},~\ref{e-pT1-normal}),
is created:
\begin{equation}
{\hat{\cal P}} (c_i) \;\; = \;\;
\left(
\begin{array}{c}
\aPair{t_{i,1}}{\pTa{t_{i,1}}}	\\
\aPair{t_{i,2}}{\pTa{t_{i,2}}}	\\
	\vdots			\\
\aPair{t_{i,j}}{\pTa{t_{i,j}}}	\\
	\vdots
\end{array}
\right)
  \label{e-list-tp}
\end{equation}

\noindent
Every tag $t_{ij}$ in ${\hat{\cal P}}$ is compared to
the most likely tag $t_{i,m}$ in ${\hat{\cal P}}$.
If the ratio of their probabilities is below a threshold $\tau\,$,
$t_{ij}$ is removed from ${\hat{\cal P}}$~:
\begin{equation}
\frac{\pTa{t_{ij}}}{\pTa{t_{i,m}}} \;\; < \;\; \tau
\end{equation}

\noindent
Removing less likely tags leads to a reduced list ${\hat{\cal P}}_r(c_i)$
that is then split into a reduced tag list $\hat t_r(c_i)$
and a reduced probability vector $\vec p_r(c_i)$
that jointly describe a reduced ambiguity class $r_y\,$.
From among all predefined $r_i$
(cf. e.g. Eq.~\ref{e-ri-probvec-exm}),
we select the one that has
the same tag list $\hat t(r_i)$ as the ``ideal'' reduced class $r_y$
and the most similar probability vector $\vec p(r_i)$
according to the cosine measure.
This $r_i$ is considered to be the most likely among all predefined $r_i$
in the context of both the current $c_i$
and the previous $r_{i-1}$.  \\

$T_2$ will map a sequence of $r_i$, from right to left,
to a sequence of tags $t_i$.
Therefore,
the construction of $T_2$ requires estimating
the most likely $t_i$
in the context of both the current $r_i$
and the following $t_{i+1}$.
To determine this $t_i$, a probability $\pTb{t_{ij}}$
is estimated for every tag $t_{ij}$ of the current $r_i\,$.
In the final position,
$\pTb{t_{ij}}$ depends on the current $r_i$
and on the following sentence boundary $\#_{i+1}\,$:
\begin{equation}
\pTb{t_{ij}} = p (t_{ij} | r_i\; \#_{i+1})
\approx \frac{p (t_{ij} | \#_{i+1}) \cdot p (t_{ij} | r_i)}
		{p (t_{ij})}
  \label{e-pT2-init}
\end{equation}

\noindent
In another than the final position,
$\pTb{t_{ij}}$ depends on the current $r_i$
and the following tag $t_{i+1}\,$:
\begin{equation}
\pTb{t_{ij}} = p (t_{ij} | r_i\; t_{i+1})
\approx \frac{p (t_{ij} | t_{i+1}) \cdot p (t_{ij} | r_i)}
		{p (t_{ij})}
  \label{e-pT2-normal}
\end{equation}

\noindent
The latter $p(t_{ij})$, $p(t_{ij}|t_{i+1})$, and $p(t_{ij}|\#_{i+1})$
are estimated from the training corpus, and
$p(t_{ij}|r_i)$ is extracted from the probability vector $\vec p(r_i)\,$.

The $t_i$ with the highest probability $\pTb{t_i}$
is the most likely tag
in the context of both the current $r_i$ and the following $t_{i+1}$
(Eq.s~\ref{e-pT2-init},~\ref{e-pT2-normal}).

\section{Construction of the FSTs \label{s-construct}}

The construction of $T_1$ is preceded by defining all $c_i$ and $r_i$,
and estimating their contextual probabilities.
In this process,
all words in the training corpus,
that are initially annotated with POS tags,
are in addition annotated with ambiguity classes $c_i$.

\begin{figure}[ht]
\vspace{1ex}
\begin{center}
\begin{minipage}[b]{44mm}
(a)\spc{-2ex}
\includegraphics[scale=0.45,angle=0]{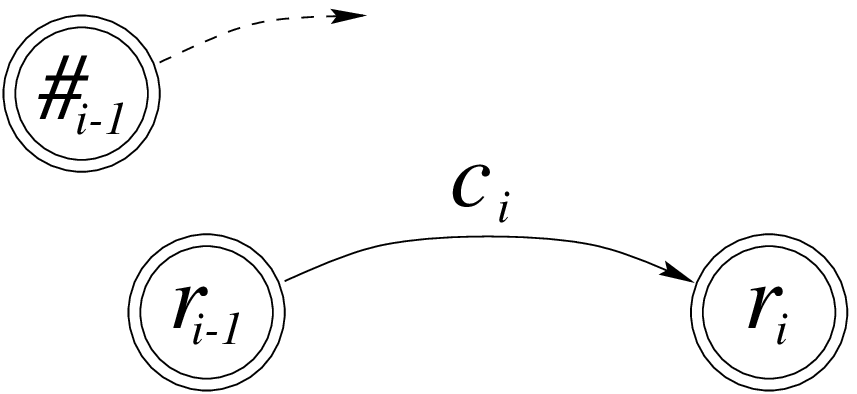}
\end{minipage}
\spc{10ex}
\begin{minipage}[b]{44mm}
(b)\spc{1.2ex}
\includegraphics[scale=0.45,angle=0]{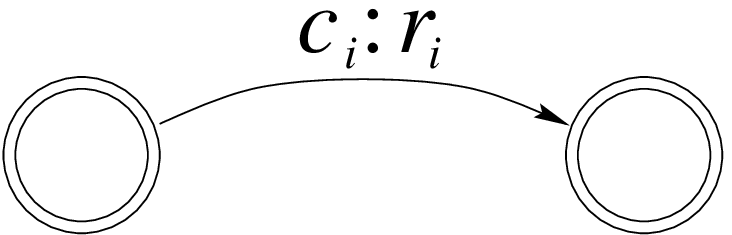}
\end{minipage}

\vspace{0.5ex}
\caption{
Two stages in the construction of $T_1$
	\label{f-T1}}
\end{center}
\end{figure}

In $T_1$,
one state is created for every $r_i$ (output symbol),
and is labeled with this $r_i$
(Fig.~\ref{f-T1}a).
An initial state, not corresponding to any $r_i$,
is created in addition.
From every state,
one outgoing arc is created for every $c_i$ (input symbol),
and is labeled with this $c_i$.
The destination of every arc is the state of the most likely $r_i$
in the context of both the current $c_i$ (arc label)
and the preceding $r_{i-1}$ (source state label)
which is estimated as described above.
All arc labels are then changed from simple symbols $c_i$
to symbol pairs $c_i${\tt:}$r_i$ (mapping $c_i$ to $r_i$)
that consist of the original arc label and the destination state label.
All state labels are removed
(Fig.~\ref{f-T1}b).
Those $r_i$ that are unlikely in any context disappear from $T_1$
because the corresponding states have no incomming arcs.
$T_1$ accepts any sequence of $c_i$
and maps it, from left to right,
to the sequence of the most likely $r_i$ in the given left context.

The construction of $T_2$ is preceded by
annotating the training corpus in addition with
reduced ambiguity classes $r_i$,
by means of $T_1$.
The probability vectors $\vec p(r_i)$ of all $r_i$
are then re-estimated.
The contextual probabilities of tags,
are estimated only at this point
(Eq.s~\ref{e-pT2-init},~\ref{e-pT2-normal}).

\begin{figure}[ht]
\vspace{1ex}
\begin{center}
\begin{minipage}[b]{44mm}
\includegraphics[scale=0.45,angle=0]{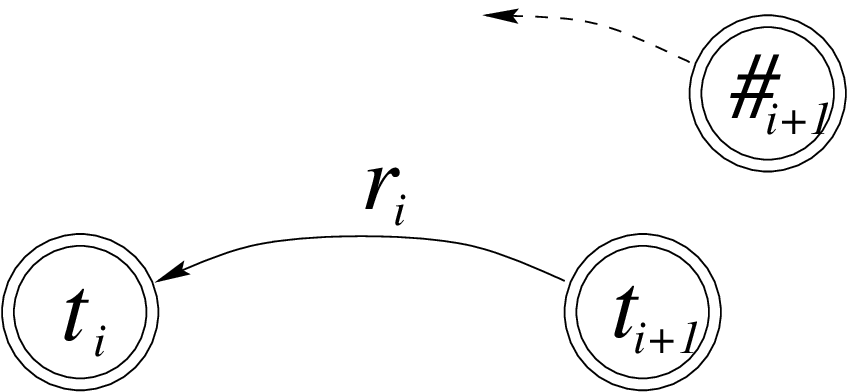}
\spc{-2ex}(a)
\end{minipage}
\spc{10ex}
\begin{minipage}[b]{44mm}
\includegraphics[scale=0.45,angle=0]{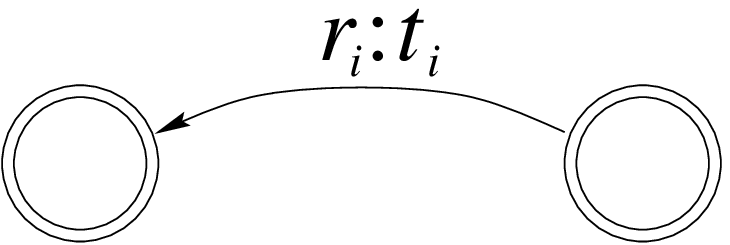}
\spc{1.2ex}(b)
\end{minipage}

\vspace{0.5ex}
\caption{
Two stages in the construction of $T_2$
	\label{f-T2}}
\end{center}
\vspace{-1.5ex}
\end{figure}

In $T_2$,
one state is created for every $t_i$ (output symbol),
and is labeled with this $t_i$
(Fig.~\ref{f-T2}a).
An initial state is added.
From every state, one outgoing arc is created for every $r_i$ (input symbol)
that occurs in the output language of $T_1$,
and is labeled with this $r_i$.
The destination of every arc is the state of the most likely $t_i$
in the context of both the current $r_i$ (arc label)
and the following $t_{i+1}$ (source state label)
which is estimated as described above.
Note, this is the following tag, rather than the preceding,
because $T_2$ will be applied from right to left.
All arc labels are then changed into symbol pairs $r_i${\tt:}$t_i$
and all state labels are removed
(Fig.~\ref{f-T2}b),
as was done in $T_1$.
$T_2$ accepts any sequence of $r_i$, generated by $T_1$,
and maps it, from right to left,
to the sequence of the most likely $t_i$ in the given right context.

Both $T_1$ and $T_2$ are sequential.
They can be minimized with standard algorithms.
Once $T_1$ and $T_2$ are built,
the probabilities of all $t_i$, $r_i$, and $c_i$
are of no further use.
Probabilities do not explicitly occur in the FSTs,
and are not directly used at run time.
They are, however,
``reflected'' by the structure of the FSTs.

\section{Application of the FSTs \label{s-applic}}

Our FST tagger uses the above described $T_1$ and $T_2$,
a class-based lexicon,
and possibly a guesser to predict the ambiguity classes of unknown words
(possibly based on their suffixes).
The lexicon and guesser are also sequential FSTs,
and map any word that they accept
to a single symbol $c_i$ representing an ambiguity class
(Fig.~\ref{f-sequences}).
If a word cannot be found in the lexicon, it is analyzed by the guesser.
If this does not provide an analysis either,
the word is assigned a special $c_i$ for unknown words
that is estimated from the {\it m} most frequent tags
of all words that occur only once in the training corpus.

The sequence of the $c_i$ of all words of one sentence
is the input to our FST cascade
(Fig.~\ref{f-sequences}).
It is mapped by $T_1$, from left to right,
to a sequence of reduced ambiguity classes $r_i$.
Intuitively,
$T_1$ eliminates the less likely tags from $c_i$,
thus creating $r_i$.
Finally,
$T_2$ maps the sequence of $r_i$, from right to left,
to an output sequence of single POS tags $t_i$.
Intuitively,
$T_2$ selects the most likely $t_i$ from every $r_i\,$.

\section{Results \label{s-results}}

We compared our FST tagger on English, German, and Spanish
with a commercially available (foreign) HMM tagger
(Table~\ref{t-spd-acc-sz}).
The comparison was made on the same non-overlapping training and test corpora
for both taggers
(Table~\ref{t-corpora}).
The FST tagger was on average 10 times as fast
but slightly less accurate than the HMM tagger
(45~600 words/sec and 96.97\% versus 4~360 words/sec and 97.43\%).
In some applications such as Information Retrieval
a significant speed increase could be worth the small loss in accuracy.

\begin{table}[p]  
\vspc{0.5ex}
\begin{center}
\begin{math}
\begin{tabular}{|p{18ex}|p{8ex}|*{3}{r}|r|}  \cline{3-6}
\multicolumn{2}{l|}{~}
  & \tabhdr{7ex}{English} & \tabhdr{7ex}{German} & \tabhdr{7ex}{Spanish}
  & \tabhdr{9ex}{Average}  \\ \hline
Speed ~(words/sec)			& $T_1\!+\!T_2$
  & 47~600 & 42~200 & 46~900 & 45~600  \\
		 			& HMM
  &  4~110 &  3~620 &  5~360 &  4~360  \\ \hline
Accuracy ~(\%)				& $T_1\!+\!T_2$
  & 96.54 & 96.79 & 97.05 & 96.97  \\
					& HMM
  & 96.80 & 97.55 & 97.95 & 97.43  \\ \hline
\end{tabular}
\end{math}

\begin{center}
{\it Computer: SUN Workstation, Ultra2, with 1 CPU}
\end{center}

\vspace{-3.5ex}
\caption{
Processing speed and accuracy of the FST and the HMM taggers
	\label{t-spd-acc-sz}}
\end{center}
\end{table}

\begin{table}[p]  
\begin{center}
\begin{math}
\begin{tabular}{|p{28.5ex}|*{3}{r}|r|}  \cline{2-5}
\multicolumn{1}{l|}{~}
  & \tabhdr{7ex}{English}
  & \tabhdr{7ex}{German}
  & \tabhdr{7ex}{Spanish}
  & \tabhdr{9ex}{Average}  \\ \hline
\# States
  &        615 &        496 &       353 &        488  \\
\# Arcs
  &    209~000 &    197~000 &    96~000 &    167~000  \\ \hline
\# Tags
  &  76 &  67 &  56 &   66 \\
\# Ambiguity classes
  & 349 & 448 & 265 &  354 \\
\# Reduced ambiguity classes
  & 724 & 732 & 465 &  640 \\ \hline
\end{tabular}
\end{math}

\begin{minipage}{120mm}
\caption{
Sizes of the FST cascades and their alphabets
	\label{t-corpora}}
\end{minipage}
\end{center}
\end{table}

\begin{table}[p]  
\begin{center}
\begin{math}
\begin{tabular}{|p{28.5ex}|*{3}{r}|r|}  \cline{2-5}
\multicolumn{1}{l|}{~}
  & \tabhdr{7ex}{English}
  & \tabhdr{7ex}{German}
  & \tabhdr{7ex}{Spanish}
  & \tabhdr{9ex}{Average}  \\ \hline
Training corpus size ~(words)
  &  20~000 &  91~000 &  16~000 &   42~000 \\
Test corpus size ~(words)
  &  20~000 &  40~000 &  15~000 &   25~000 \\ \hline
\end{tabular}
\end{math}

\begin{minipage}{120mm}
\caption{
Sizes of the training and test corpora
	\label{t-corpora}}
\end{minipage}
\end{center}
\end{table}

\pagebreak

\end{document}